# Can Euroscepticism Contribute to a European Public Sphere? The Europeanization of Media Discourses about Euroscepticism across Six Countries

Anamaria Dutceac Segesten (Lund University)
Michael Bossetta (University of Copenhagen)

## Abstract

This study compares the media discourses about Euroscepticism in 2014 across six countries (United Kingdom, Ireland, France, Spain, Sweden, and Denmark). We assess the extent to which the mass media's reporting of Euroscepticism indicates the Europeanization of public spheres. Using a mixed-methods approach combining LDA topic modeling and qualitative coding, we find that approximately 70 per cent of print articles mentioning 'Euroscepticism' or 'Eurosceptic' are framed in a non-domestic (i.e. European) context. In five of the six cases studied, articles exhibiting a European context are strikingly similar in content, with the British case as the exception. However, coverage of British Euroscepticism drives Europeanization in other Member States. Bivariate logistic regressions further reveal three macro-level structural variables that significantly correlate with a Europeanized media discourse: newspaper type (tabloid or broadsheet), presence of a strong Eurosceptic party, and relationship to the EU budget (net contributor or receiver of EU funds).

## Introduction

The 2014 European Parliament (EP) elections resulted in over one quarter of the available seats going to Eurosceptic party members – the highest number to date (Treib, 2014). As a result, national media outlets in several Member States reported about the 'earthquake', 'virus', or 'rising tide' of Euroscepticism sweeping across the continent. Since research has long shown that the media's coverage of the EU influences public opinion (De Vreese, 2007), scholars are rightly turning their attention to studying

Euroscepticism in the news (Caiani and Guerra, 2017). However, prior research has not yet examined how Euroscepticism is reported by the media as a news topic. We address this missing component by focusing our analysis on the national media discourse about Euroscepticism, which comprises explicit mentions of Euroscepticism as a phenomenon and mentions of political actors who the media labels as 'Eurosceptic'.

In particular, the study seeks to test whether the national media discourses about Euroscepticism are Europeanized and further, to uncover which structural factors drive Europeanization processes in the media. Given that Euroscepticism is a topic of relevance for the entire EU polity, the media discourse about Euroscepticism may be Europeanized. A discourse is Europeanized when an issue involving non-national or EU level actors is discussed in a similar fashion within different Member States, thus allowing for "cross-border understanding and communication" (Risse, 2014, p. 11) across national publics. Europeanized discourses increase the saliency of the EU within national public spheres and enable transnational communication between them. Although Euroscepticism implies opposition to the EU and European integration, a Europeanized media discourse about Euroscepticism may facilitate mutual understanding and foster cross-border dialogue among Europeans. In order to test this argument, we ask the following research questions:

1) *To what extent are national media discourses about Euroscepticism Europeanized?*

2) *What factors explain the presence of a Europeanized media discourse about Euroscepticism?*

To answer these questions, the study employs a two-phase, mixed-methods research design. In the first phase, we select mainstream media articles published in 2014 that contain the words 'Eurosceptic' or 'Euroscepticism' across six countries: the United Kingdom, Ireland, Spain, France, Denmark, and Sweden. Using a combination of LDA topic modeling and qualitative coding, we identify the content of the articles (i.e. what topics are discussed) as well as their scope (i.e. whether the topics concern national or European issues). In the second phase, we use logistic regressions to test which structural factors motivate a Europeanized media discourse about Euroscepticism.

The findings from the first phase demonstrate that the topics associated with Euroscepticism are remarkably similar across five of the six cases studied, with the UK being the exception. Moreover, both the topic models and human coding reveal that 70 per cent of the 1,545 articles included in our dataset are European in scope. Therefore, we argue that the mainstream media discourse about Euroscepticism is, in fact, Europeanized. Although the UK is an exception, we also find that Britain served as a driver for Europeanization in the media discourse of other Member States; each of these cases cover British politics in the context of Euroscepticism.

Answering our second research question, the results of the log regressions expound three variables significantly correlated to the media's portrayal of Euroscepticism in a European context: newspaper type (tabloid or broadsheet), domestic Eurosceptic party success (loser or winner of the European Parliament elections), and relationship to the EU budget (net contributor or receiver of EU funds). Respectively, these variables correspond to structural features in the domestic media system, political system, and economic system, and we find that the media variable is the strongest predictor for Europeanization. Taken together, the findings suggest that Euroscepticism is a news topic present across national contexts and as such, can be one of the factors contributing to the emergence of a European public sphere. However, national context matters: the extent of Europeanization is moderated by structural factors in the media, party, and financial systems of EU Member States.

## Euroscepticism in the media: A new perspective

Recent literature highlights the enduring difficulties in defining and delimiting Euroscepticism (Leruth *et al.,* 2017). This draws into question the analytical utility of Euroscepticism as a theoretical framework. The distinction between 'hard' and 'soft' Euroscepticism (Taggart, 1998) only partially addresses this criticism. Szczerbiak and Taggart (2017, p. 16) acknowledge the 'extremely broad' scope of soft Euroscepticism, which may lead to wrongly-categorizing parties that are pro-integration. Furthermore, studies that attempt to operationalize Euroscepticism via typologies typically result in large 'diffuse' categories (de Wilde *et al,* 2014; Szczerbiak and Taggart, 2017, p. 14).

The difficulties in measuring Euroscepticism derive from the complexity of the EU polity and the multifaceted character of European integration. Given that these complexities are unlikely to abate, we consider scholarly efforts to categorize the varieties of Euroscepticism a noble task – but ultimately an unfruitful one. Instead of taking Euroscepticism as a dependent or independent variable, we use it here as a test case to measure the Europeanization of public spheres. Therefore, we are not interested in studying Euroscepticism as such; rather, we want to explain why the media discourse about Euroscepticism varies across national contexts.

The present study builds on previous research arguing for the influence of macro-level structural conditions in explaining attitudes towards the EU (Koehler *et al.,* 2018; De Vries, 2018). Since 'attitudes towards the EU are framed by the national circumstances in which people live and their evaluations of these conditions' (De Vries, 2018, p. 3), public opinion towards the EU is intimately linked to the configuration of domestic media, political, and economic systems. For example, Hobolt and De Vries (2016) argue that national identity, political partisanship, and economic interest are the primary influences on public opinion when it comes to the evaluation of the European integration process (Hobolt and De Vries, 2016).

Citizens' perceptions of these three factors are heavily influenced by national press. Media coverage about the EU has been shown to affect public opinion towards the EU's policies (Maier and Rittberger, 2008) and overall performance (De Vreese, 2007). The mass media, to paraphrase Trenz (2013, p. 35), can be seen as either facilitators or obstacles of European integration. Media actors contribute to the "communicative construction of Europe" (Hepp *et al.*, 2016, p. 13) through the representation of common spaces shared by Europeans, regardless of their national community. The media's influence extends beyond their national readership, as topics chosen and highlighted by the press are likely to be diffused further through face-to-face conversations or via platform-mediated communication on social media (Bossetta, Dutceac Segesten, and Trenz, 2017).

However, media reporting about the EU is not uniform (De la Porte and Van Dalen, 2018; Fracasso *et al.*, 2015) and can be attributed to differences in how media operate across the continent. National media differ in terms of their structural development (Hallin and Mancini, 2004) and the established norms of communication between politicians and journalists (Pfetsch, 2013). Thus, when EU issues are 'downloaded' (Börzel, 2002, p. 196) to national discourse arenas via the press, media reporting about the EU often varies across media systems and more pointedly, across national contexts.

At the same time, national media exhibit some commonalities in their communication practices, and rigorous empirical tests demonstrate that these similarities can be distinguished along four major geographic regions in (western) Europe: Northern, Central, Western, and Southern (Brüggemann *et al.*, 2014). Media coverage about the EU can therefore 'converge' along certain topics, especially during Europe-wide events such as the European Parliament elections (de Wilde, *et al.* 2014). Topic convergence across national media underpins the development of a public space for shared political discussion. In the European context, this process has been described as the 'Europeanization of public spheres' (Risse, 2014).

## The Europeanization of media discourse

While the concept of Europeanization typically refers to the degree of congruence between national and European institutions (Börzel and Risse, 2007), we define Europeanization broadly as any process whereby a feature of the domestic (whether it be an identity, a policy, or a discourse) takes on a European dimension. Furthermore, our discursive approach to Europeanization aligns with what Trenz (2008, p. 278) describes as 'discursive interaction', whereby the 'Europeanization of public and media communication can be analyzed as a process that enlarges the scope of public discourse beyond the territorial nation state'. Following this argument, a media discourse can be considered Europeanized when the same politicians and events from around Europe are featured prominently in domestic news coverage.

Europeanization can be either a vertical process, when the media foregrounds pan-European or supranational issues or actors, or a horizontal one, when the media covers an event or actor specific to another Member State (Brüggemann and Kleinen von Königslöw, 2007, pp. 21-24). If the media discourse about an issue is both Europeanized and discussed in a similar manner across national contexts, cross-border dialogue between those media outlets (and their readerships) can occur. Such instances of converged media coverage across Member States would indicate the Europeanization of public spheres (Risse, 2014).

Existing research shows that Europeanization of news coverage takes place, albeit to varying degrees. Examining the various media representations of Europe longitudinally from 1982-2013, Hepp *et al*. (2016) observe a slow progression towards a transnational coverage of EU politics but with a national tinge, leading to a segmented European public sphere. De la Porte and Van Dalen (2016), too, find that EU's socio-economic strategy is Europeanized across media in four countries. Fracasso *et al.* (2015) identify a pan-European coverage of austerity and the euro crisis across all Member States but with national variations, depending on the relevance of the topic for domestic audiences.

Focusing on Euroscepticism, Barth and Bijsmans (2018) find a convergence of news frames and negative evaluation of the EU in German and British broadsheets around the Maastricht Treaty. Dutceac Segesten and Bossetta (2017), meanwhile, show that journalists' and citizens' discussions of Euroscepticism can exhibit higher degrees of Europeanization than other EU-related issues in both print and social media. Even if their study was limited to two countries (Sweden and Denmark), it highlighted the potential for Euroscepticism to catalyze Europeanization, which has yet to be tested across media systems.

## Research Design

The present study's research design achieves precisely such a comparison across media systems. We select media discourses about Euroscepticism in six countries that vary along several dimensions: size, region, media system, success of a national Eurosceptic party, and relationship to the EU budget. More specifically, we focus on three criteria that relate to structural features of domestic media, political, and economic systems. The selection is designed to stratify both similarity and difference across our cases. Regarding media systems, we select two countries from each of Hallin and Mancini's (2004) three types. From the democratic corporatist system, we choose Denmark and Sweden; from the polarized pluralist model, France and Spain; and from the liberal model, the United Kingdom and Ireland. Within each system, one country contains a Eurosceptic party that won the 2014 EP elections (DK, FR, UK), and the other contains a Eurosceptic party that had only little or moderate success (SE, ES, IE). Finally, in terms of macroeconomic features, Ireland and Spain are net receivers of EU funds, whereas the other four countries (SE, DK, FR, UK) are net contributors to the EU budget.

For each national case, we include two types of print media: quality newspapers and tabloids. We then compare the main topics presented in media articles containing the word 'Euroscepticism' and/or 'Eurosceptic' across 2014. During a European Parliament election year, the chance that Europe-related issues will appear in the media increases (Grill and Boomgaarden, 2017). Thus, by choosing the year 2014, we purposely oversample for the presence of Europeanized topics. Should the Eurosceptic discourse not be Europeanized under these circumstances, the Europeanization phenomenon would be less likely to occur outside of election years.

Our aim is to assess whether the topics reported in these articles are comparable in terms of their content (a necessary aspect for Europeanization to occur) and their context (whether the topics are presented in a national or European scope). If topics and contexts converge across national media, we can speak of a Europeanized discourse. We then seek to uncover the structural mechanisms underpinning those instances where evidence for Europeanization is found. Grounded in prior research, we test three variables (relating to the media, political, and economic system) that we hypothesize to influence the scope of coverage about Euroscepticism.

The first variable is **newspaper type**, and we distinguish between broadsheet and tabloid formats. Pfetsch *et al.* (2008, p. 31) argue that broadsheets are 'more inclined to stress European dimensions' than tabloid press. Even in terms of the amount of coverage, quality newspapers tend to have more space reserved to European issues than tabloids (De Vreese et al., 2006).

The second variable is the **success of Eurosceptic political actors** in each domestic case. National identity and party cues are two of the primary determinants of public attitudes towards the EU (Hobolt and De Vries, 2016, pp. 421-22); Eurosceptic parties therefore act as double influencers, since they mobilize national attachment *and* send party cues. According to Boomgaarden and de Vreese (2013), political elites with an anti-EU leaning increase the visibility of the EU in national news during EP campaigns. In addition, the coverage of EP elections in each Member State tends to give priority to nationally relevant topics and politicians (Schuck et al., 2012).

Our third variable, the role of **economic factors,** has been investigated by Lubbers and Scheepers (2010, p. 787), who find that '[g]rowing media attention increases political Euroscepticism in countries with a negative EU budget balance, whereas it decreases such skepticism in countries with a positive balance'. Also examining representations of the economic dimension in the media, De Wilde (2012) argues that the politicization of the EU budget is connected to the framing of debates as horizontal conflicts among different EU Member States – not as vertical conflicts between EU Member States and the European Commission. Similarly, Koehler *et al*. (2018) demonstrate that macroeconomic indicators such as low GDP growth, when coupled with media representations of the EU as a solution for economic crises, predict support for European integration.

Based on this previous research, we formulate the following hypotheses:

**H1**. The tabloids' Eurosceptic discourse will be more national in scope than the broadsheets'.
**H2**. The more successful the Eurosceptic party, the more national the scope of the discourse about Eurosceptics will be.
**H3**. Newspapers from net budget contributor countries will have a more national discourse about the Eurosceptics than net receivers.

## Method and data

We identified the media discourse about Euroscepticism by searching for the keywords 'Eurosceptic' and 'Euroscepticism' (in the respective languages[1]) across three media article databases: LexisNexis, Retriever, and Infomedia. We chose these criteria after including other words and phrases indicative of Euroscepticism (e.g., 'anti-EU', 'critical of the EU', 'opposed to European integration'), but we found that the addition of these words did not significantly increase the size of our sample. Important to emphasize is that not all of the articles were centrally *about* Euroscepticism as a phenomenon; rather, the article merely needed to include one of the two words to be included in the study. We embrace this broad selection criterion precisely because we want to chart the variety of contexts alongside which Euroscepticism is mentioned, to cast a wide net for topic convergence. This approach, versus including keywords for various issues related to Euroscepticism like immigration or austerity, also ensures the selection is strictly comparable across cases.

Our sample includes all articles from the top two broadsheets and top two tabloids[2] by circulation in each country during the period January 1 - December 31, 2014. This resulted in 1,545 articles, distributed as follows: Spain (467), Denmark (440), Sweden (275), France (207), United Kingdom (129), and Ireland (26). The difference in the amount of articles shows that Eurosceptic discourses have an uneven presence across national contexts. Ireland is an outlier with the lowest number of articles, an exception that is likely due to two factors. The first is the lack of availability of Irish newspapers in the media databases (the three newspapers included were all the Irish print media that they offered access to). The second is the economy of scale in publishing, with British newspapers readily available for Irish readers.

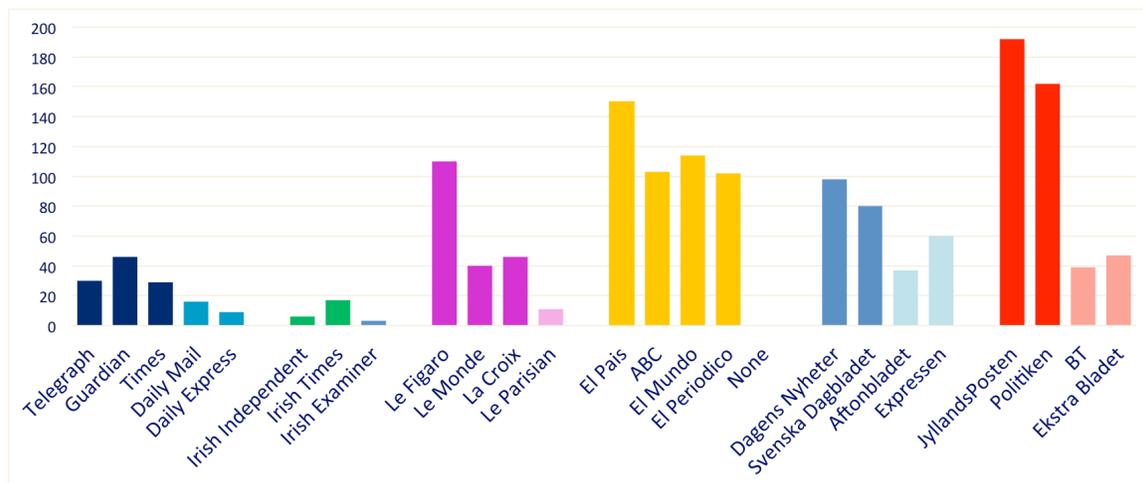

Figure 1: Number of Articles per Newspaper

Figure 1 depicts the distribution of articles by country, media outlet, and newspaper type (i.e., broadsheet or tabloid). For each country, tabloids are depicted as lightly shaded bars. Some countries have a predominance of broadsheet newspapers whereas others see a strong presence of tabloids,[3] in particular Britain and Sweden (McLachlan and Golding, 2000; Andersson 2013). To implement the first stage of our design, we investigated the specific ways in which newspapers cover Euroscepticism. We extracted the predominant topics associated with our keywords for each national media environment using unsupervised LDA topic modeling for articles from each national case. LDA topic modeling, a computational content analysis method based on patterns of word co-occurrence across documents, has been used to identify the print media coverage of policy domains such as the government support for the arts (DiMaggio *et al.*, 2013) or nuclear technology (Jacobi *et al.*, 2016). One of the advantages of topic modeling is that it enables comparative analysis even when data distribution across cases is uneven, allowing us to analyze the Irish case despite the low number of articles collected. The results of the model provide a strict, statistically-derived measure to compare the generated topics across cases.

Before fitting each national media corpus as a model, the text was subjected to pre-processing using the 'quanteda' (Benoit *et al.,* 2017) package for the programming software R. To increase the accuracy of our models, we stemmed and removed stopwords for each national case using quanteda's default language dictionaries, and we subsequently added stopwords deemed irrelevant to our analysis. This process was conducted through repeated iterations of automated content analysis of the top 50 words for each national case, until all 50 top words were deemed relevant for study. For each respective language, we removed commonly appearing words that were deemed not useful in conveying political content as well as the first names of prominent politicians (e.g. 'Angela', 'David', 'Jean-Claude'). Since some outlets referred to politicians by their first and last names, and other outlets preferred only the last name, we dropped the first name to reduce noise. In a similar vein, we also combined the names of political parties and institutions, treating them as one word ('uni-grams'). These pre-processing steps were taken in order to increase the coherence and accuracy of generated topics.

The LDA modeling was performed using the 'lda' package for the programming software R (Chang, 2015). While no standardized rules exist for selecting the number of topics, based on the size of the data and several iterative tests to maximize coherency, we chose 10 clusters for Ireland, 15 for the UK, and 20 each for Denmark, Sweden, Spain, and France.[4] We set the model's parameters to 1500 iterations with an alpha of .1. We then qualitatively assigned labels to each of the topics according to their content (see Appendix B). If a topic could not be inferred, we discarded it from the analysis. Then, we manually categorized each topic as either 'National' or 'European', depending on the whether the topic signaled content relevant to domestic or non-national politics.

To conduct the bivariate logistic regressions, we first used human coding to identify the primary context of each article as either National or European (the inter-coder reliability of a 60-article sample was 83 per cent). An article was coded 'National' if the main actors in the article were national actors and the relevance of the reported events was placed in a domestic context. For example, if a Swedish newspaper talked about Jimmie Åkesson (the leader of the Sweden Democrats) and the coverage was mostly about the agenda of the party for the European elections, the article would be coded 'National'. The 'European' code was assigned when non-national actors were foregrounded, and/or when events were given relevance beyond the specific national community. That was the case if, for example, a Danish newspaper referred to the leader of the Sweden Democrats and talked about their expected chances of winning a seat in the European Parliament. Articles that were more general opinion pieces about the meaning of Euroscepticism about European identity, the future of the European Union, or the threat of absenteeism across the 28 Member States were also coded 'European'. Both the horizontal and the vertical dimensions of Europeanization, as outlined by Brüggemann and Kleinen von Königslöw (2007), were subsumed as 'European' in our coding.

## Results

Having read and coded all of the media articles included in the study, we found that the LDA topic models performed well despite differences in language and corpus size. In the countries where 20 topics were requested (Spain, Denmark, Sweden, and France), the LDA model generated 18, 17, 17, and 16 coherent topics, respectively. The British topic model produced 13 interpretable topics out of 15, and the Irish case 7 out of 10. After manually labeling the coherent topics by scope – i.e. whether they concerned national or European politics – we report the results below in Table 1 as percentages to account for the differences in corpus size.

| Country | National Topics | European Topics | Incoherent Topics |
|---|---|---|---|
| United Kingdom | 60% | 25% | 15% |
| Ireland | 0% | 70% | 30% |
| France | 20% | 60% | 20% |
| Spain | 15% | 75% | 10% |
| Sweden | 30% | 55% | 15% |
| Denmark | 25% | 60% | 15% |

Table 1: Scope of Topics Generated by LDA Modeling (%)

National topics generally referred to contestation between national parties, either in the context of the EP or local elections, as well as articles on domestic issues like the state of economy (for the full list of topics see Appendix B). In every case but the UK, Euroscepticism was proportionately discussed more often in a European context, with Irish media being exceptional in discussing Euroscepticism solely in a European context. In aggregate numbers, 61 out of the 88 topics (or 70 per cent) identified by the topic modeling were European in scope.

Surprisingly, the *content* of the European topics across countries was remarkably similar across cases. Figure 2 is a visualization of a directed network graph of all the European topics, with color-coded edges marking the connection between national media and the generated topics. The size of the nodes corresponds to the total amount of times a topic appears in our model. Fracasso *et al.* (2015) applied a similar method in their study of the transnational coverage of the euro crisis. However, our method differs in that the nodes are topics and not, like in their case, country mentions.

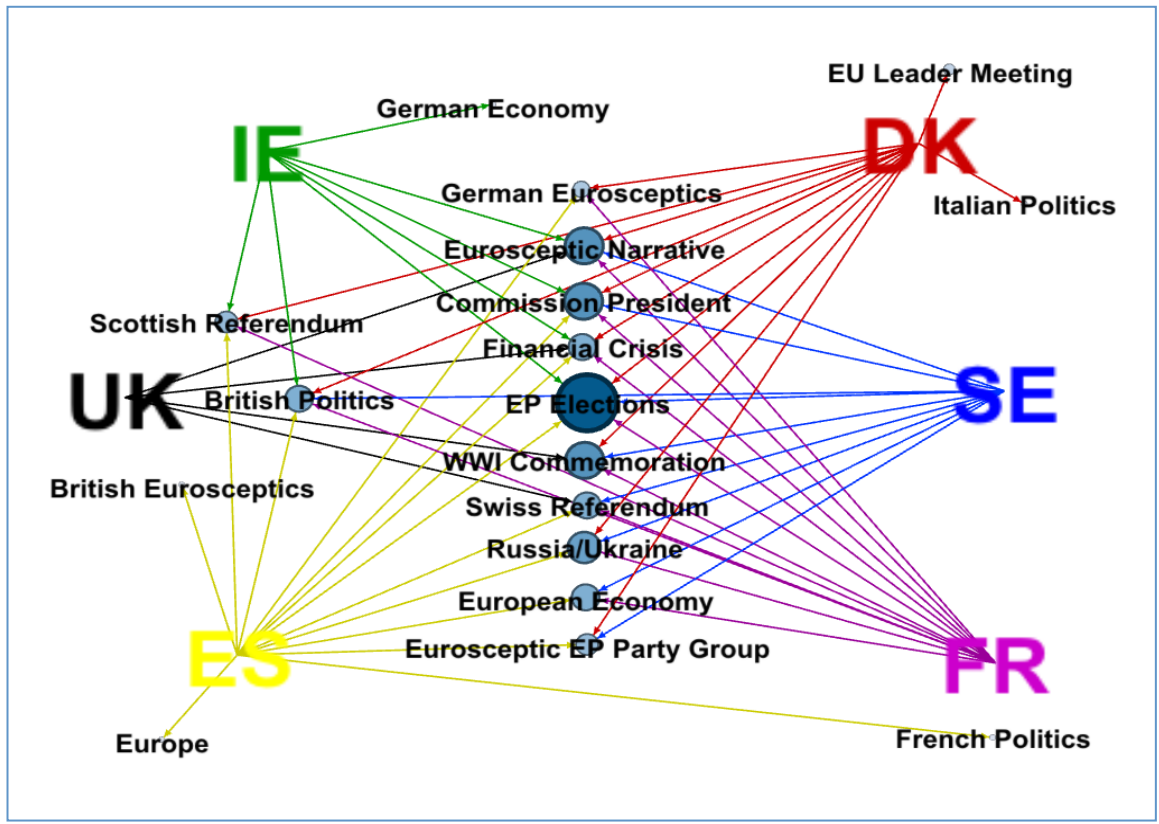

**Figure 2: Network Graph of European Context Topics**

Topics in the center of the graph were present in three or more of the countries' media. Starting from the top, *German Eurosceptics* refers to coverage of the German Eurosceptic party Alternative für Deutschland (AfD). *Eurosceptic Narrative* refers to reportage about the 'rise of Euroscepticism' across Europe following the 2014 EP elections. *Commission President* refers to the nomination of the European Commission President, a newly instated process that consisted of public debates involving candidates from European political parties (i.e. Spitzenkandidaten). *Financial Crisis* refers to topics relating to the Eurozone crisis, while *EP Elections* is a category assigned to general reportage about the elections. *WWI Commemoration* refers to a series of events where European leaders met to commemorate the 100[th] anniversary of the start of World War I. *Swiss Referendum* refers to a public consultation held in February 2014 on instating quotas on migrants from the EU. *Russia/Ukraine* relates to coverage of the Euromaidan revolution as well as the Russian annexation of Crimea. *European Economy* refers to general coverage of economic affairs in Europe, while *Eurosceptic EP Party Group* is a topic dedicated to the attempted formation of a Eurosceptic party group by national party leaders Nigel Farage, Marine Le Pen, and Geert Wilders.

At the left side of Figure 2 are nodes specifically related to *British Politics*. Every country had a topic dedicated to politics in the UK, generally focused around Prime Minister David Cameron, the Euroscepticism of his center-right Conservative Party, and/or the promised referendum on Britain's EU membership. All countries but Sweden generated a topic relating to the *Scottish Referendum* to leave the United Kingdom in September 2014. Lastly, some European topics are not shared by the other cases. Spain, for example, has isolated topics on *French Politics, British Eurosceptics* (i.e., the UK Independence Party), as well as *Europe*, referring to more general reflections about European identity and the future of the EU. Ireland has an isolated topic dealing specifically with the *German Economy*, while Denmark's isolated topics concern *Italian Politics* as well as an *EU Council Meeting* that was held in Denmark.

Figure 2 also shows that relatively few topics are unlinked (6 of 88, or 7 per cent), which signals that most topics of European scope are shared, common stories across our national cases. This seems to suggest that when either 'Eurosceptic' or 'Euroscepticism' appears in national media reportage, 'similar frames are used in the various public arenas so as to allow cross-border understanding and communication' (Risse, 2014, p. 11). Bolstering the picture presented by the graph and answering our first research question (*To what extent are national media discourses about*

*Euroscepticism Europeanized?)*, our qualitative coding of the 1,545 articles resulted in 1075 (or 70 per cent) being labeled as European in scope. This human coding of the articles' context helps validate the LDA models, where our labeling of generated topics also rendered 70 per cent as European in scope.

## Structural factors in the Europeanization of Euroscepticism discourse

To answer our second research question – *what factors explain the presence of a Europeanized media discourse about Euroscepticism?* – we tested which of our three independent variables is most closely associated with the presence of a European scope in our data. We did not differentiate by country since we consider our selection process to represent enough variation across cases. Moreover, we do not consider the small number of articles collected for the Irish case to affect the overall quality of our results, with the potential exception of the last bivariate regression on the EU budget relationship (where more cases are needed to increase confidence in the generalizability of the results).

To answer this second research question, we first performed a chi-square test of the strength of association between European topics and our three structural variables: newspaper type, position of national Eurosceptic party, and relationship to EU budget. In all three cases the p-value was significant at p<.05, allowing us to proceed with performing bivariate log regressions. We calculated the odds ratio, a method typically used for categorical variables such as ours, in order to assess the likelihood that articles with a European scope would appear in broadsheets as opposed to tabloids (Table 2). As expected, the odds ratio registers in favor of broadsheets hosting articles with a European scope by a multiple higher than three, thus confirming **H1.**

|  | Rate | Risk Ratio | Odds | Odds Ratio | Log odds ratio |
|---|---|---|---|---|---|
| Broadsheet | 0.7294 |  | 2.6852 |  |  |
|  |  | 1.7376 |  | 3.7257 | 1.3153 |
| Tabloid | 0.4198 |  | 0.7234 |  |  |

Table 2: Odds Ratio for Newspaper Type

Moving to our second hypothesis, we calculated the likelihood that a European scope corresponds to articles from countries where the Eurosceptics won the 2014 EP elections (UK, FR, DK), as opposed to countries where Eurosceptics had little to moderate success (IE, ES, SE). In line with hypothesis **H2**, Table 3 shows that content presented in a European context was over one and a half times more likely to occur in newspapers from countries where Eurosceptics were electorally weak, compared to where they performed strongly.

|  | Rate | Risk Ratio | Odds | Odds Ratio | Log odds ratio |
|---|---|---|---|---|---|
| Non-Eurosceptic | 0.7448 |  | 2.9184 |  |  |
|  |  | 1.1467 |  | 1.575 | 0.4543 |
| Eurosceptic | 0.6495 |  | 1.8529 |  |  |

Table 3: Odds Ratio for the Presence of Winning Eurosceptic Party

Table 4 shows that newspapers from net receiving countries (IE, ES) were over two and a half times more likely to publish content with a European scope than net contributors (UK, FR, DK, SE). This finding supports our final hypothesis, **H3**.

|  | Rate | Risk Ratio | Odds | Odds Ratio | Log odds ratio |
|---|---|---|---|---|---|
| Receiver | 0.8316 |  | 4.9398 |  |  |
|  |  | 1.3124 |  | 2.8556 | 1.0493 |
| Giver | 0.6337 |  | 1.7299 |  |  |

Table 4: Odds Ratio for EU Budget Relationship

While all three of the variables tested had a significant effect on whether media discourse relating to Euroscepticism was framed in a national or European context, the strength of association differed across variables. According to our analysis, the strongest predictor for European coverage is newspaper type, with broadsheets being over three times more likely to cover Eurosceptic actors or events in a European (i.e., non-national) context. The second strongest predictor of a European context was the Member State's relation to the EU budget (with an odds ratio of 2.85 in favor of states that receive more than they contribute), and the weakest was the degree of domestic Eurosceptic party success (the European context was 1.57 times more likely to be present when there was no strong Eurosceptic party on the national scene).

## Discussion

Taken together, the results of the content analysis suggest that 70 per cent of the media discourse about Euroscepticism included in our study is Europeanized. More specifically, the LDA topic modeling reveals that many of the issues relating to Euroscepticism appear in several of the cases – despite differences in their media, political, and economic systems. In other words, the Europeanization of media coverage about Euroscepticism is a phenomenon that occurs despite national variation. Although we did not make any distinction between vertical and horizontal Europeanization *a priori*, our analysis reveals both types of Europeanization were present in domestic news reporting. Vertical Europeanization (mentions of supra-national institutions, policies, and processes such as the creation of a European Parliament party group of Eurosceptic MEPs) was to be expected in the context of 2014 EP elections. The presence of horizontal Europeanization (comparisons across EU Member States such as the birth of a German Eurosceptic party or the austerity problems in Southern Europe) further allows us to claim that Europeans have access to similar frames of reference regarding Euroscepticism. Driven by the media, such possibilities for *common understanding* satisfy one of the primary conditions for the existence of a European public sphere as outlined by Risse (2014).

The horizontal Europeanization identified, however, was not uniform. Our study corroborates the idea that some countries elicit more references in western European media than others. With the exception of Poland and Hungary, East European Member States were entirely absent from our dataset. By contrast, the UK is the predominant reference node for the reportage on Euroscepticism, although the pan-European narrative of the 'rise of the Eurosceptics' is also an important common trope. The media's preoccupation with the UK supports existing research that some countries have more net gravity in attracting foreign news attention than others. Both Koopmans and Statham (2010, p. 67) and Fracasso *et al*. (2015, p.7) find that France, Germany, and the UK account for half of the claims in foreign media coverage about the EU. While the other five countries in our study report about British politics (e.g., the Scottish referendum or the referendum on the EU membership of the UK), the British reader was hard pressed to find any media references to events taking place in continental Europe. This discrepancy in coverage may indicate, but may have also contributed to, the perception of isolation and idiosyncrasy that is often associated with the UK in its relationship to Europe. Our findings help contextualize the British media's coverage of the EU before Brexit, where the print media is argued to have heavily shaped the national debate (Seaton, 2016).

In addition, our analysis finds that media type is the strongest predictor for Eurosceptic coverage to take a European scope in national press. Confirming the results of Pfetsch (2008, p. 31) and Hepp *et al.* (2016, p. 106), our data show tabloids were more national than European in their coverage of news, a finding that may be explained by the media's proclivity to appeal to a domestic readership. In comparison with broadsheet readers, tabloid readers tend to have lower social status (Chan and Goldthorpe, 2007) and be less educated – two socio-economic indicators that are often attributed to supporters of Eurosceptic parties (Lubbers and Scheepers, 2010, p. 812). The effect of these individuals' exposure to mostly national coverage of pan-European phenomena, such as the rise of Euroscepticism, may further reduce their sense of shared belonging to a European public.

Broadsheets were much more likely to place Euroscepticism in a European context, perhaps due to their more informed, affluent readership that also has a higher tendency to support European integration (Serrichio *et al.,* 2012). France and Spain are interesting cases to highlight, since their media landscape does not include many tabloid newspapers. French and Spanish broadsheets are the ones to set the standard of reporting about Euroscepticism, and they do so by exposing readers to stories that tend to be more European than national in scope. In Denmark, tabloids are not numerous; consequently, the European scope of Eurosceptic news coverage is widespread -- despite the fact that Denmark is less integrated into the polity (e.g. maintaining a national currency and several opt-outs of EU treaties). Thus, our analysis suggests that newspaper type matters more than the degree of a Member State's integration, contradicting the findings of Pfetsch *et al*. (2008) and Hepp *et al.* (2016).

Moreover, our results indicate that the presence of an electorally strong Eurosceptic party increases the odds of a national scope of news and opinion pieces. This finding is rather unsurprising, as journalists must cover the homegrown Eurosceptic phenomenon for their domestic audience. The presence of popular domestic Eurosceptic parties leads to a more national coverage of Euroscepticism, which may spread or exacerbate anti-EU attitudes among that national public (Van der Brug, 2016, p. 269). Interestingly, though, we also find that countries without a winning Eurosceptic party in the EP elections still dedicated space to the coverage of Euroscepticism occurring elsewhere. The Spanish and Swedish media discussed Euroscepticism predominantly as a pan-European phenomenon (the size of the media coverage of 'Eurosceptic' and 'Euroscepticism' in Spain and Sweden is comparable to that of France and the UK), and they problematized the consequences of such a development for the EU. Such reportage serves as a prime example of how Euroscepticism is a topic that can drive the Europeanization of national media discourses.

Regarding our third variable that examined the country's relationship to the EU budget, previous research has argued that media coverage in countries with a positive EU budget balance decreases Euroscepticism (Lubbers and Scheepers, 2010, p. 787). Supporting this argument, our results suggest that net receivers from the EU budget place Euroscepticism in a European context more often than net contributors. Therefore, media in Member States benefitting from EU funds seem more likely to externalize the Eurosceptic phenomenon than report about domestic Eurosceptic manifestations. Journalists in these Member States, perhaps taking their cues from the national political elites, prefer to not politicize EU matters domestically, or even, like Bobba and Seddone (2017) observe in the case of Italy, give a more favorable coverage to EU matters than to national issues.

## Conclusion

This study sought to answer whether the national media discourse relating to Euroscepticism is Europeanized. We conclude that it is, and that this Europeanization process is driven by national media outlets. Through expounding the most salient issues discussed in news articles containing the words 'Eurosceptic' and 'Euroscepticism' during 2014, we find a strong convergence in topics in five of the six cases studied. While the discourse about Euroscepticism in the UK appears the least Europeanized, we find that British Euroscepticism nevertheless drives *other* Member States' media to discuss Euroscepticism under the shared frame of British politics. We did not test direct cross-border communication initiatives between Member State media or citizens; however, we consider the similar topics discussed by the media in the context of reporting about Euroscepticism sufficient to fulfill the precondition of 'mutual understanding' (Risse, 2014, p. 11) needed for a European public sphere to emerge.

The results of our bivariate logistic regressions provide firm ground to assert the importance of domestic structural factors for Europeanization processes. Variation in the degree of Europeanization of media discourse – at least in the case of Euroscepticism - is explained by media, party, and economic features in domestic Member State contexts. These findings confirm the importance of nationally-specific structural factors in explaining not only Euroscepticism (De Vries, 2018), but also the Europeanization of public spheres.

Our study has several limitations. The media databases archive an incomplete selection of media sources, which did not allow us to consistently include the top two tabloids and broadsheets by circulation per country. Moreover, in choosing to focus on a European election year, we may have disproportionately weighted the presence of Europeanized topics. Future research could compare election with non-election year coverage, to test if our three structural factors still hold their explanatory power outside of electoral cycles. In addition, we encourage scholars to use the upcoming 2019 European Parliament elections to examine more nuanced frame convergence (Barth and Bijsmans, 2018) in topics relating to Euroscepticism.

Furthermore, when defining our selection criteria, we used keywords that explicitly referenced Euroscepticism as a concept; it is possible that we missed some articles relating to the phenomenon that did not use our keywords explicitly. In other words, national variations in journalists' phrasing of 'Euroscepticism' may bias our results. Lastly, we invite other scholars to extend the comparison to other countries in order to strengthen the model proposed here. We did not include Germany in this study because of strict copyright restrictions on media, and we could not accurately represent Member States from Central or Eastern Europe due to language barriers.

Despite these limitations, our findings support the assertion that Euroscepticism is a common topic of discussion in media discourses across Member States and that the variations in the presence of a European rather than national scope depend on systemic features of domestic media, politics, and the economy. Therefore, we conclude that: the discourse about Euroscepticism is Europeanized; this Europeanization process is driven by the media; and is influenced by the type of newspaper, the existence of a successful Eurosceptic party, and whether a country is a net contributor or receiver of EU funds.

Van der Brug, W. (2016) 'European Elections, Euroscepticism and Support for Anti-European Union Parties.' In Van der Brug, W. and de Vreese (eds.) *(Un)intended Consequences of EU Parliamentary Elections* (Oxford: Oxford University Press), pp. 256-275.

---

[1] For a list of all keywords, see Appendix A.

[2] In some countries there were no tabloids present, in which case we included a major regional newspaper.

[3] Due to LexisNexis limitations, we were only able to include three newspapers for Ireland. For the same reason, we could not make a selection of the two leading tabloids and two quality newspapers in the British case. We chose to include five newspapers in the British case, three tabloids and two quality newspapers, in order increase the number of articles for better results in our topic models.

[4] While changing the number of clusters altered the words present in each topic, the topics remained relatively stable despite whether 10, 15, or 20 topics were generated.

# Appendix A: Keywords per country

| Country | Keywords |
|---|---|
| UK | Euroscepticism, Eurosceptic/s |
| Ireland | Euroscepticism, Eurosceptic/s |
| France | Euroscepticisme, eurosceptique/s |
| Spain | Euroescepticismo, euroesceptico/a/s |
| Sweden | EU-kritik, EU-kritisk/a |
| Denmark | EU-kritik, EU-kritisk/a |

# Appendix B: Topics per country

| France | | |
|---|---|---|
| Number | Scope | Topic |
| 1 | National | National Politics |
| 2 | National | Green Parties |
| 3 | National | Local Elections |
| 4 | National | Local Politics |
| 5 | European | EP elections |
| 6 | European | Economy |
| 7 | European | Russia/Ukraine |
| 8 | European | Financial Crisis |

| 9 | European | Commission President |
| 10 | European | Swiss Referendum |
| 11 | European | German Eurosceptics |
| 12 | European | British Politics |
| 13 | European | WWI Commemoration |
| 14 | European | Scottish Referendum |
| 15 | European | Eurosceptic Narrative |
| 16 | European | EP elections |
| 17 | X | X |
| 18 | X | X |
| 19 | X | X |
| 20 | X | X |

| Spain | | |
|---|---|---|
| Number | Scope | Topic |
| 1 | National | Catalonian Independence |
| 2 | National | National Politics |
| 3 | National | Culture/Identity |
| 4 | European | French Politics |
| 5 | European | Europe |
| 6 | European | German Eurosceptics |
| 7 | European | British Politics |
| 8 | European | Financial Crisis |
| 9 | European | Commission President |
| 10 | European | British Eurosceptics |
| 11 | European | Russia/Ukraine |
| 12 | European | Eurosceptic EP Party Group |
| 13 | European | European Economy |
| 14 | European | European Economy |
| 15 | European | Swiss Referendum |
| 16 | European | EP Elections |
| 17 | European | Scottish Referendum |
| 18 | European | EP Elections |
| 19 | X | X |
| 20 | X | X |

| Sweden | | |
|---|---|---|
| Number | Scope | Topic |
| 1 | National | National Politics |

| 2 | National | National Politics |
| 3 | National | National Politics |
| 4 | National | National Politics |
| 5 | National | National Politics |
| 6 | National | National Eurosceptics |
| 7 | European | Russia/Ukraine |
| 8 | European | Eurosceptic EP Party Group |
| 9 | European | Commission President |
| 10 | European | Swiss Referendum |
| 11 | European | Eurosceptic Narrative |
| 12 | European | EP Elections |
| 13 | European | British Politics |
| 14 | European | EP elections |
| 15 | European | European Economy |
| 16 | European | WWI Commemoration |
| 18 | X | X |
| 19 | X | X |
| 20 | X | X |

| Denmark | | |
|---|---|---|
| Number | Scope | Topic |
| 1 | National | Welfare |
| 2 | National | National Politics |
| 3 | National | National Economy |
| 4 | National | Patent Court |
| 5 | National | National Politics |
| 6 | European | Eurosceptic EP Party Group |
| 7 | European | Italian Politics |
| 8 | European | Scottish Referendum |
| 9 | European | British Politics |
| 10 | European | WWI Commemoration |
| 11 | European | Financial Crisis |
| 12 | European | German Eurosceptics |
| 13 | European | Leader Meeting |
| 14 | European | Russia/Ukraine |
| 15 | European | Commission President |
| 16 | European | Eurosceptic Narrative |
| 17 | European | EP elections |
| 18 | X | X |
| 19 | X | X |

| 20 | X | X |

| United Kingdom | | |
|---|---|---|
| Number | Scope | Topic |
| 1 | National | Farage/Clegg Debate |
| 2 | National | Cameron's Cabinet Shuffle |
| 3 | National | Boris Johnson Mayor |
| 4 | National | Scottish Referendum |
| 5 | National | Local Elections |
| 6 | National | National Eurosceptics (UKIP) |
| 7 | National | Brexit |
| 8 | National | Economy |
| 9 | National | Blair, Iraq War |
| 10 | European | Eurosceptic Narrative |
| 11 | European | WWI Commemoration |
| 12 | European | Swiss Referendum |
| 13 | European | Financial Crisis |
| 14 | X | X |
| 15 | X | X |

| Ireland | | |
|---|---|---|
| Number | Scope | Topic |
| 1 | European | German Economy |
| 2 | European | Brexit |
| 3 | European | EP elections |
| 4 | European | Eurosceptic Narrative |
| 5 | European | Scottish Referendum |
| 6 | European | Financial Crisis |
| 7 | European | Commission President |
| 8 | X | X |
| 9 | X | X |
| 10 | X | X |